\documentclass[sigconf]{acmart}
\usepackage{balance}
\usepackage{verbatim}
\usepackage{algorithm}
\usepackage{algorithmic}
\usepackage{enumerate}
\usepackage{fancyvrb}
\usepackage{balance}

\def\BibTeX{{\rm B\kern-.05em{\sc i\kern-.025em b}\kern-.08emT\kern-.1667em\lower.7ex\hbox{E}\kern-.125emX}}
\copyrightyear{2019}
\acmYear{2019}
\setcopyright{acmcopyright}
\acmConference[MM '19] {Proceedings of the 27th ACM International Conference on Multimedia}{October 21--25, 2019}{Nice, France}
\acmBooktitle{Proceedings of the 27th ACM International Conference on Multimedia (MM'19), Oct. 21--25, 2019, Nice, France}
\acmPrice{15.00}
\acmDOI{10.1145/3343031.3350928}
\acmISBN{978-1-4503-6889-6/19/10}

\begin{document}
\fancyhead{}
%
\title{Dense Feature Aggregation and  Pruning for RGBT Tracking}
\author{Yabin Zhu}
\affiliation{%
 \institution{Key Laboratory of Intelligent Computing and Signal Processing of Ministry of Education, School of Computer Science and Technology, Anhui University}
 \streetaddress{Jiu long road}
 \city{Hefei}
 \state{Anhui}
 \country{China}}
\email{zhuyabin0726@foxmail.com}

\author{Chenglong Li}
\affiliation{%
 \institution{Key Laboratory of Intelligent Computing and Signal Processing of Ministry of Education, School of Computer Science and Technology, Anhui University}
 \institution{Institute of Physical Science and Information Technology, Anhui University}
 \streetaddress{Jiu long road}
 \city{Hefei}
 \state{Anhui}
 \country{China}}
\email{lcl1314@foxmail.com}

\author{Bin Luo}
\affiliation{%
 \institution{Key Laboratory of Intelligent Computing and Signal Processing of Ministry of Education, School of Computer Science and Technology, Anhui University}
 \streetaddress{Jiu long road}
 \city{Hefei}
 \state{Anhui}
 \country{China}}
\email{luobin@ahu.edu.cn}

\author{Jin Tang}
\affiliation{%
 \institution{Key Laboratory of Intelligent Computing and Signal Processing of Ministry of Education, School of Computer Science and Technology, Anhui University}
 \institution{Key Laboratory of Industrial Image Processing and Analysis of Anhui Province}
 \streetaddress{Jiu long road}
 \city{Hefei}
 \state{Anhui}
 \country{China}}
\email{tangjin@ahu.edu.cn}

\author{Xiao Wang}
\affiliation{%
 \institution{Key Laboratory of Intelligent Computing and Signal Processing of Ministry of Education, School of Computer Science and Technology, Anhui University}
 \streetaddress{Jiu long road}
 \city{Hefei}
 \state{Anhui}
 \country{China}}
\email{wangxiaocvpr@foxmail.com}

%

%
\begin{abstract}
How to perform effective information fusion of different modalities is a core factor in boosting the performance of RGBT tracking. This paper presents a novel deep fusion algorithm based on the representations from an end-to-end trained convolutional neural network. To deploy the complementarity of features of all layers, we propose a recursive strategy to densely aggregate these features that yield  robust representations of target objects in each modality. In different modalities, we propose to prune the densely aggregated features of all modalities in a collaborative way. In a specific, we employ the operations of global average pooling and weighted random selection to perform channel scoring and selection, which could remove redundant and noisy features to achieve more robust feature representation. Experimental results on two RGBT tracking benchmark datasets suggest that our tracker achieves clear state-of-the-art against other RGB and RGBT tracking methods.

\end{abstract}
%

%

%
\maketitle

\section{Introduction}
\begin{figure}[t]
  \centering
  \includegraphics[width=\columnwidth]{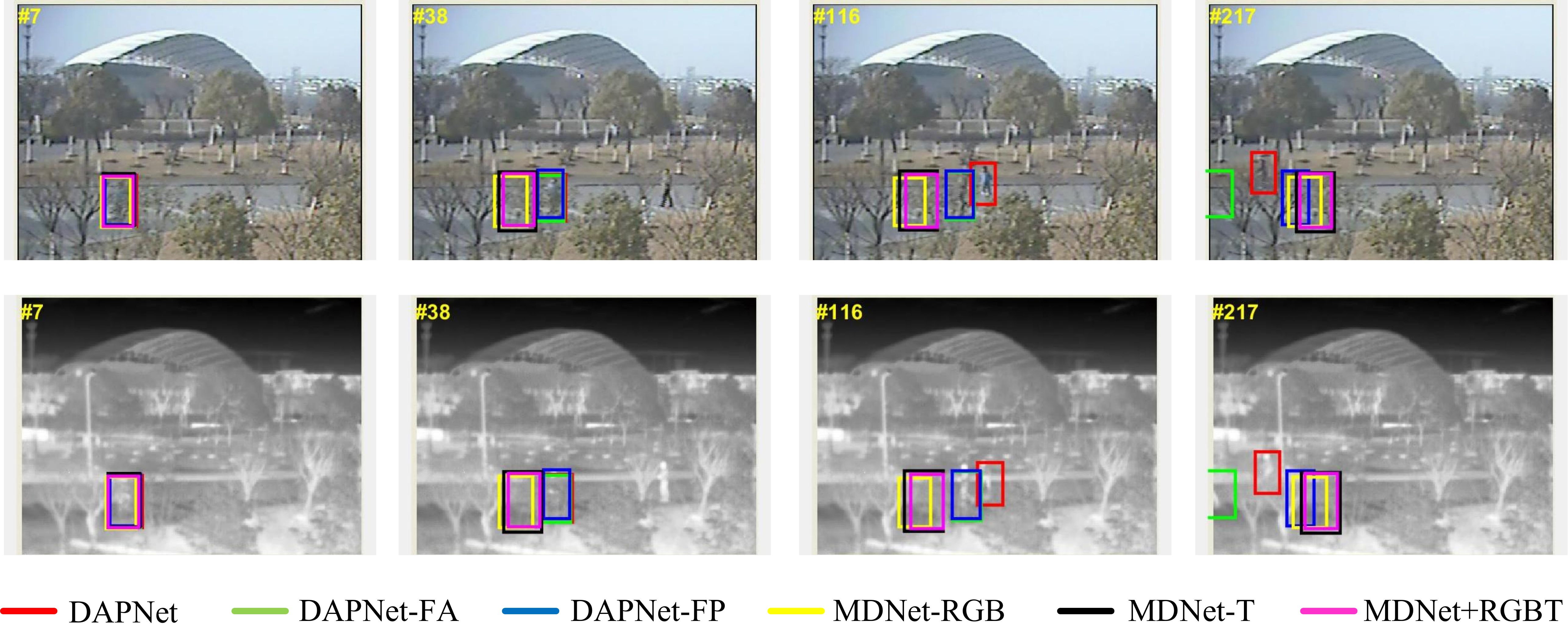} \\
  \caption{Visual examples of our tracker comparing with MDNet with different inputs including RGB, thermal and RGBT, denoting MDNet-RGB, MDNet-T and MDNet+RGBT respectively. Also, we show two variants of our DAPNet, i.e., DAPNet-FA that removes feature pruning of DAPNet and DAPNet-FP that removes dense feature aggregation in DAPNet. The results demonstrate effectiveness of dense feature aggregation and feature pruning in our DAPNet. }\label{fig::right_corner}
\end{figure}
Given the initial ground truth, the task of RGBT tracking is to track a particular instance in sequential frames using RGB and thermal infrared information.  Recently, it has received increasing attention as it is able to achieve robust tracking performance in challenging environments by utilizing inter-modal complementarity. Despite much progress in RGBT tracking ~\cite{Li18eccv,Li2018Learning,li2017regle,Li17rgbt210}, there are still many problems need to be solved, where how to effectively fuse RGB and thermal infrared sources is a core factor in boosting tracking performance and still not solved well.

Some RGBT tracking methods ~\cite{Conaire07mva,Conaire2006Comparison} used manual weights to achieve adaptive integration of RGB and thermal data, but their generality and scalability are low. Other methods ~\cite{Wu2011Multiple, Liu12infosci} performed joint sparse representation in Bayesian filtering framework by fusing features or reconstruction coefficients of different modalities. It usually introduces redundant and noisy information when some individual source is malfunction. Recently, some RGBT tracking works ~\cite{lan18aaai,Li17tsmcs} focused on introducing modality weights to achieve adaptive fusion of different source data. Lan \emph{et al.} ~\cite{lan18aaai} used the max-margin principle to optimize the modality weights according to classification scores. Li \emph{et al.} ~\cite{Li2016Learning} employed reconstruction residues to regularize modality weight learning.
However, these works would fail when the reconstruction residues or classification scores are unreliable to reflect modal reliability. The above works only rely on  handcrafted feature to localize objects and thus be difficult to handle the challenges of significant appearance changes caused by deformation, background clutter and partial occlusion and low illumination within each modality. Li \emph{et al.}~\cite{li18nuecom} adopted a two-stream CNN network and a fusion network to fuse these two modalities, but they only adapted highly semantic features which are unable to locate targets precisely.

In this paper, we propose a novel approach, namely Dense feature Aggregation and  Pruning Network (DAPNet), for RGBT tracking. Shallow features could encode appearance and spatial details of targets and thus are beneficial to achieving precise target localizations, while deep features are more effective to capture target semantics which can effectively identify the target category. Some existing works~\cite{Tao2016HyperNet,Li2018SiamRPN} usually employ specific feature layers for sparse feature aggregation to enhance tracking performance. To make best use of deep features, our method instead recursively aggregates features of all layers in a dense fashion.  As shown in Fig.~\ref{fig::FACS}, our DAPNet makes full use of shallow-to-deep spatial and semantic features to achieve more accurate tracking results. In addition, we also compress feature channels to reduce redundancy and use the max pooling operation to transform different sizes of feature maps into the same scale. To reduce network parameters and capture common properties of different modalities, we make RGB and thermal backbone networks sharing same parameters.

The aggregated RGBT features are noisy and redundant as some of them are useless or even interfering in locating a certain target. That is to say, only a few convolutional filters are active and a large portion of ones contain redundancy and irrelevant information in describing a certain target, which leads to over-fitting, as demonstrated in ~\cite{Li2019Target,Li2016Learning}.  To handle these problems, we propose a collaborative feature pruning method to remove noisy and redundant feature maps for more robust tracking. Existing works ~\cite{He2017Channel,luo2017thinet} exploited reconstruction-based methods, which seek to do channel pruning by minimizing the reconstruction error of feature maps between the pruned model and a pre-trained model. However, these methods incorrectly preserve the actual redundant channels by minimizing the reconstruction errors of the feature maps. In this paper, we apply the idea of channel pruning to solve our problem, and improve it with just simple operations to achieve excellent tracking performance. 

In a specific, the feature pruning module is followed by the feature aggregation module, but it does not increase the number of network parameters and is only deployed in training phase with slight computational cost. The feature pruning module consists of two steps, \emph{i.e,} channel scoring and channel selection which are realized by a global average pooling and weighted random selection. Through this feature pruning method, we choose to inactivate some of the feature channels in each iteration of training, resulting in a more robust convolutional feature representation. Once the training is done, the network parameters of aggregation are fixed and feature pruning are removed during online tracking.

We validate the effectiveness of the proposed tracking framework on two RGBT tracking benchmark datasets, including GTOT ~\cite{Li16tip} and RGBT234 ~\cite{li2019rgb}. We summarize our major contributions as follows. First, we propose a novel end-to-end trained deep network for accurate RGBT tracking. By deploying all enhanced deep features, our tracker is able to handle the challenges of significant appearance changes caused by partial occlusion, deformation and adverse environmental conditions, etc. Extensive experiments show that the proposed method outperforms other state-of-the-art trackers on RGBT tracking datasets. Second, we propose a dense feature aggregation module to recursively integrate features of all layers into a same feature space. Finally, to further eliminate effects of noisy features after dense feature aggregation, we design a feature pruning module to obtain more robust feature representation and achieve better tracking performance.

\section{Related Work}
In this section, we give a brief review of tracking methods closely related to this work.

\subsection{RGBT Tracking}
RGBT tracking receives more and more attention in the computer vision community with the popularity of thermal infrared sensors. 
Recent methods on RGBT tracking mainly focus on sparse representation because of its capability of suppressing noise and errors ~\cite{Wu2011Multiple}, ~\cite{Li2016Learning}. 
Wu \emph{et al.} ~\cite{Wu2011Multiple} concatenate the image patches from RGB and thermal sources into a one-dimensional vector that is then sparsely represented in the target template space.  
Collaborative sparse representation based trackers is proposed by Li \emph{et al.} ~\cite{Li2016Learning} to jointly optimize the sparse codes and modality weights online for more reliable tracking. 
And Li \emph{et al.}~\cite{Li18eccv} further consider heterogeneous property between different modalities and noise effects of initial seeds in the cross-modal ranking model. 
These methods rely on handcrafted features to track objects, and thus are difficult to handle the challenges of significant appearance changes caused by background clutter, occlusion, deformation within each modality.

\subsection{Feature Aggregation for Tracking} 
Feature aggregation  ~\cite{Yu2018Deep, AntonioarXiv} is becoming more and more popular to improve network performance by enhancing the representation of features. Without exception, in the field of visual tracking ~\cite{li18nuecom,HDT16cvpr,ECO17cvpr,danelljan2016beyond}, there are many methods to improve tracking performance by the skill of feature aggregation. 
Li \emph{et al.}~\cite{li18nuecom} design a FusionNet to directly aggregate RGB and thermal feature maps from the outputs of two-stream ConvNet. 
A aggregation of handcrafted low-level and hierarchical deep features is proposed by Danelljan \emph{et al.}~\cite{danelljan2016beyond,ECO17cvpr} by employing an implicit interpolation model to pose the learning problem in the continuous spatial domain, which enable efficient integration of multi-resolution feature maps.  
Qi \emph{et al.}~\cite{HDT16cvpr} take full advantages of features from different CNN layers and used an adaptive Hedge method to hedge several CNN trackers into a stronger one. 
Li \emph{et al.}~\cite{Li2018SiamRPN} propose a new architecture to aggregate the middle to the deep layer features, which not only improves the accuracy but also reduces the model size. 
Different from these methods, we proposed a novel feature aggregation and pruning framework for RGBT tracking, which recursively aggregates all layer deep features while compressing feature channels.

\begin{figure*}[t]
  \centering
  \includegraphics[width=\textwidth]{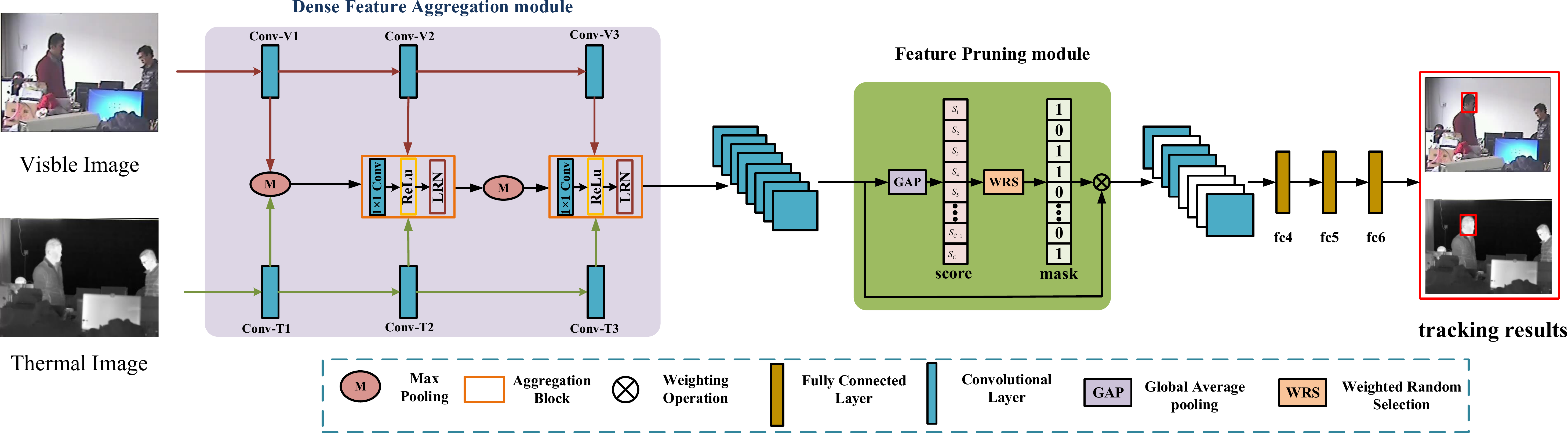} \\
  \caption{Diagram of the proposed network architecture.  The network consists of two major modules, i.e., a dense feature aggregation module and a feature pruning module.}\label{fig::FACS}
\end{figure*}

\section{Proposed Approach}
In this section, we first describe the proposed training framework for RGBT object tracking. 
Then we introduce two modules of dense feature aggregation and feature pruning in detail.

\subsection{ Network Architecture} 
As shown in Figure ~\ref{fig::FACS}, the proposed network consists of a fully convolutional dense feature aggregation module, a feature pruning module and three fully connected layers(fc4-6) for binary classification. 
Following MDNet~\cite{Nam2015Learning}, we choose a lighter VGG-M network ~\cite{vgg15iclr}(conv1-3) as our backbone. 
Herein, two modalities adopt the same backbone network and share parameters. 
Different from original MDNet, max pooling layer is removed after conv2 layer, and dilated convolution with the rate as 3 is applied for extracting a dense feature map with a higher spatial resolution. 
Then we aggregate all features of two modalities by a feature aggregation block. 
To reduce feature noise interference, the aggregated features are selected by a feature pruning module. 
Finally, the optimized features are classified by three fully connected layers and softmax cross-entropy loss. 
The network has $K$ branches (i.e. the $K$ domains) which are denoted by the last fully connected layers, in other words, training sequences {$fc6^{1}-fc6^{K}$}. 
More details of multi-domain learning can be found in ~\cite{Nam2015Learning}. 

\subsection{Dense Feature Aggregation}
The dense feature aggregation module is a feature fusion strategy that aims at strengthening the fully convolutional architecture through parallel hierarchical structure. 
Its task is to better process and propagate the features from the original network to the classifier. 
Aggregation block, the main building blocks of the dense features aggregation module, learns to combine the output of multiple convolutional layers, and extract all spatial and semantic information from shallow to deep features. 
We implement the feature aggregation module as a parallel feature processing branch that can be plugged into any CNN architecture. 
Our aggregation consists of a stacking sequence of aggregation blocks, each one iteratively combining the output from the backbone and from the previous aggregation block, as shown in Figure ~\ref{fig::FACS}. 
When the inputs of a aggregation block have different scales, we use the max pooling operation to keep their sizes consistent. 
The proposed aggregation block are implemented as a ${\bf 1x1}$ convolution followed by non-linear(ReLU) and normalization(LRN) operations, as shown in Figure~\ref{fig::FACS}. 
Our structure can be combined with any existing pre-trained models without disrupting the propagation of the original features. 
To fully fuse features of different modalities, the outputs of RGB and thermal backbone network are all connected to the aggregation block. 
To map the features of the two modalities to the same feature space, we select the same backbone network and share the parameters. 
The dense feature aggregation module aggregates the spatial and semantic information of the two modalities from shallow to deep, and compresses the feature channels so that more rich and effective feature representation can be obtained. 
Let $B(\cdot)$ denote the aggregation operation of the feature, and we have
\begin{equation}
\label{eq::RFA}
B({\bf x}_1,...,{\bf x}_n) =LRN(\sigma (\sum_i{\bf W}_i {\bf x}_i + {\bf b})),~i=1,2,...,n.
\end{equation}
where $\sigma$ is the non-linear activation, and ${\bf W}_i$ and ${\bf b}$ are the weights and bias in the convolution respectively. Local Response Normalization (LRN) is a normalized function, and ${\bf x}$ denotes the input of a aggregation block.

\subsection{Feature Pruning}
To eliminate noisy and redundant information introduced by the dense feature aggregation and inspired by ~\cite{Hou2019WCD}, we propose a feature pruning mechanism. 
Note that our motivation is significantly different from ~\cite{Hou2019WCD}. %
\cite{Hou2019WCD} is a dropout technique to avoid overfitting, while we aim to prune out the redundant and noise features and retain the most discriminative features to a certain target for more effective localization, in addition to avoiding overfitting of network training. 
By this way, the learning of the effective feature representations is enhanced, and useless features are suppressed.
In a specific, channel dropout ~\cite{Hou2019WCD} is used between two adjacent convolution layers, and the operations of Global Average Pooling (GAP), Weighted Random Selection (WRS) and Random Number Generation (RNG) are used to select some channels to achieve regularization. 
While our feature pruning operations will choose some channels with greater impact on target localization. 
If directly using ~\cite{Hou2019WCD} to achieve our goal, it would make our network difficult to optimize due to the dense aggregation structure of our network.
In addition, the RNG operation would increase the randomness of feature selection and thus we leave out it.

Lin \emph{et al.} ~\cite{Min2013Network} propose to replace the fully connected layer (FC) with GAP to solve the problem of overfitting and excessive FC layer parameters in convolutional neural networks. 
Zhou \emph{et al.} ~\cite{Zhou2016Learning} reveal that using GAP can make convolutional neural networks have excellent localization ability. 
Therefore, in this paper, we use GAP to obtain the activation state of each feature channel,
\begin{equation}
\label{eq::GAP}
score_{c} =\frac{1}{{ W}\times{ H}} \sum_{j=1}^{W}\sum_{k=1}^{H} {x}_c(j,k)
\end{equation}
where $W$ and $H$ are the width and height of feature map. ${x}_c$ denotes the feature map of the $c$-th channel.

In this paper, we do not directly use the $score$ to perform channel selection, but instead adopt WRS ~\cite{Efraimidis2006Weighted} which is a more efficient algorithm. 
Specifically, each channel $x_c$ has $score_c$, a random number $r_{c}\in (0,1)$ is generated. The key value $key_c$ is computed as 
\begin{equation}
\label{eq::WRS}
key_{c}= r_{c}^{\frac{1}{score_{c}}}
\end{equation}

The $M$ items of the largest key values are selected, where $M=N*wrs\_ratio$, $N$ is channels number and $wrs\_ratio$ is a parameter indicating how many channels are selected after ${\bf WRS}$. 
The detailed steps of the whole collaborative channel pruning can be found in Algorithm~\ref{alg:Framwork}.

\begin{algorithm}[htb] 
\caption{Feature Pruning.} 
\label{alg:Framwork} 
\begin{algorithmic}[1] 
\REQUIRE ~~\\ 
feature channel $x_c$, channel selection rate $wrs\_ ratio$;
\ENSURE ~~\\ 
\STATE Calculate channel $score_c$ by Eq.~\ref{eq::GAP}; 

\STATE For each c, $r_c=random(0,1)$ and calculate $key_{c}$ value by Eq.~\ref{eq::WRS}; 

\STATE Select the $M=N*wrs\_ratio$ items with the largest $key_c$; 

\STATE Obtain the feature channels after feature pruning;

\RETURN $\tilde{ x}_i$; 
\end{algorithmic}
\end{algorithm}

\subsection{Network Training}
In this section, we describe the training details of our network. 
First, we initialize the parameters of the first three convolutional layers using the pre-trained model of the VGG-M network ~\cite{vgg15iclr}. 
While fully connected layers are initialized randomly. 
Then, we train the whole network by the Stochastic Gradient Descent (SGD) algorithm, where each domain is handled exclusively in each iteration. 
In each iteration, mini-batch is constructed from 8 frames which are randomly chosen in each video sequence. 
And we draw 32 positive and 96 negative samples from each frame which results in 256 positive and 768 negative data altogether in a mini-batch. 
The samples whose the IoU overlap ratios with the ground truth bounding box are larger than 0.7 are treated as positive, and the negative samples have less than 0.5. 
For multi-domain learning with $K$ training sequences, we train the network with 100 epoch iterations by softmax cross-entropy loss. 
We train our network using 77 video sequences randomly selected form RGBT234 dataset ~\cite{li2019rgb} and test it on GTOT dataset ~\cite{Li16tip}. 
For another experiment, we train our network on all 50 video sequences from GTOT dataset and test it on RGBT234 dataset.

\subsection{Tracker Details}
In tracking, the $K$ branches of domain-specific layers (the last fc layer) are replaced with a single branch for each test sequence. 
Moreover, the feature pruning module is removed. 
During the tracking process and online fine-tuning, we fix the convolutional filters ${w_1,w_2,w_3}$ and fine-tune the fully connected layers ${w_4,w_5,w_6}$ because the convolutional layers would have generic tracking information whereas the fully-connected layers have the video-specific knowledge.  
Given the first frame pair with the ground truth of target object, we draw 500 positive (IoU with ground truth is larger than 0.7) and 5000 negative samples (IoU with ground truth is smaller than 0.5), and train the new branch with 10 iterations. 
Given the $t$-th frame, we draw a set of candidates $\{{\bf z}_t^i\}$ from a Gaussian distribution of the previous tracking result ${\bf z}^*_{t-1}$, where the mean of Gaussian function is set to ${\bf z}^*_{t-1}=(a_{t-1},b_{t-1},s_{t-1})$ and the covariance is set as a diagonal matrix $diag\{0.09r^2,0.09r^2,0.25\}$. 
$(a,b)$ and $s$ indicate the location and scale respectively and $r$ is the mean of $(a_{t-1},b_{t-1})$. For the $i$-th candidate ${\bf z}_t^i$, we compute its positive and negative scores using the trained network as $f^+({\bf z}_t^i)$ and $f^-({\bf z}_t^i)$, respectively. 
The target location of the current frame is the candidate with the maximum positive score as:
\begin{equation}
\label{edge:score}
{\bf z}_t^* = \arg\max_{{\bf z}_t^i} f^+({\bf z}_t^i),~i=1,2,...,N,
\end{equation}
where $N$ is the number of candidates. 
We also apply bounding box regression technique~\cite{MDNet15cvpr} to improve target localization accuracy. 
The bounding box regressor is
trained only in the first frame to avoid potential unreliability of other frames. 
If the estimated target state is sufficiently reliable, \emph{i.e.} $f^+({\bf z}_t^*)>0.5$, we adjust the target locations using the regression model. 
More details can be referred to~\cite{MDNet15cvpr}.

\begin{table*}[t]\footnotesize 
\setlength{\belowcaptionskip}{0.3cm}
\caption{Attribute-based PR/SR scores (\%) on RGBT234 dataset against with eight RGBT trackers. The best and second results are in \textcolor{red}{red} and \textcolor{green}{green} colors, respectively.}
\centering
\begin{tabular}{ c | c  c  c  c  c  c  c  c | c }
	\hline
   & SOWP+RGBT &CFNet+RGBT &KCF+RGBT &L1-PF &CSR-DCF+RGBT & MEEM+RGBT & SGT   &MDNet+RGBT1 & DAPNet \\\hline
 NO &{86.8}/53.7 & 76.4/56.3 & 57.1/37.1 & 56.5/37.9 &82.6/{60.0} & 74.1/47.4 &\textcolor{green}{87.7}/55.5 & 86.2/\textcolor{green}{61.1} & \textcolor{red}{90.0}/\textcolor{red}{64.4}  \\
 PO & 74.7/48.4 & 59.7/41.7 & 52.6/34.4 & 47.5/31.4 & 73.7/\textcolor{green}{52.2} & 68.3/42.9 & \textcolor{green}{77.9}/51.3 & 76.1/51.8 & \textcolor{red}{82.1}/\textcolor{red}{57.4} \\

 HO & 57.0/37.9 & 41.7/29.0 & 35.6/23.9 & 33.2/22.2 &59.3/40.9 & 54.0/34.9 & 59.2/39.4 & \textcolor{green}{61.9}/\textcolor{green}{42.1} & \textcolor{red}{66.0}/\textcolor{red}{45.7} \\

 LI & \textcolor{green}{72.3}/46.8 & 52.3/36.9 &51.8/34.0 & 40.1/26.0 & 69.1/\textcolor{green}{47.4} & 67.1/42.1 & 70.5/46.2 & 67.0/45.5 & \textcolor{red}{77.5}/\textcolor{red}{53.0} \\

 LR & 72.5/46.2 & 55.1/36.5 & 49.2/31.3 & 46.9/27.4 & 72.0/47.6 & 60.8/37.3 & \textcolor{green}{75.1}/47.6 &\textcolor{red}{75.9}/\textcolor{red}{51.5} & 75.0/\textcolor{green}{51.0} \\

 TC & 70.1/44.2 & 45.7/32.7 & 38.7/25.0 &37.5/23.8 & 66.8/46.2 & 61.2/40.8 & \textcolor{green}{76.0}/47.0 & 75.6/\textcolor{green}{51.7} & \textcolor{red}{76.8}/\textcolor{red}{54.3}  \\

 DEF & 65.0/46.0 & 52.3/36.7 & 41.0/29.6 & 36.4/24.4 & 63.0/46.2 & 61.7/41.3 & \textcolor{green}{68.5}/\textcolor{green}{47.4} & 66.8/47.3 & \textcolor{red}{71.7}/\textcolor{red}{51.8} \\

 FM & 63.7/38.7 & 37.6/25.0 & 37.9/22.3 & 32.0/19.6 & 52.9/35.8 & 59.7/36.5 & \textcolor{red}{67.7}/\textcolor{green}{40.2} &58.6/36.3 &\textcolor{green} {67.0}/\textcolor{red}{44.3} \\

SV &66.4/40.4 & 59.8/43.3 & 44.1/28.7 & 45.5/30.6 & 70.7/49.9 & 61.6/37.6 & 69.2/43.4 &\textcolor{green}{73.5}/\textcolor{green}{50.5} & \textcolor{red}{78.0}/\textcolor{red}{54.2} \\

 MB & 63.9/42.1 & 35.7/27.1 & 32.3/22.1 & 28.6/20.6 & 58.0/42.5 & 55.1/36.7 & 64.7/43.6 & \textcolor{red}{65.4}/\textcolor{green}{46.3} &\textcolor{green}{65.3}/\textcolor{red}{46.7} \\

 CM & 65.2/43.0 & 41.7/31.8 & 40.1/27.8 & 31.6/22.5 & 61.1/44.5 & 58.5/38.3 & \textcolor{green}{66.7}/45.2 & 64.0/\textcolor{green}{45.4} & \textcolor{red}{66.8}/\textcolor{red}{47.4} \\

 BC & 64.7/41.9 & 46.3/30.8 & 42.9/27.5 & 34.2/22.0 & 61.8/41.0 & 62.9/38.3 & \textcolor{green}{65.8}/41.8& 64.4/\textcolor{green}{43.2} &\textcolor{red}{71.7}/\textcolor{red}{48.4}\\\hline

 ALL &69.6/45.1 & 55.1/39.0 &46.3/30.5 & 43.l1/28.7 & 69.5/49.0 & 63.6/40.5 & 72.0/47.2 & \textcolor{green}{72.2}/\textcolor{green}{49.5} & \textcolor{red}{76.6}/\textcolor{red}{53.7} \\\hline
\end{tabular}
\label{tb::AttributeResults}
\end{table*}

\begin{figure*}[t]
  \centering
  \includegraphics[width=\textwidth]{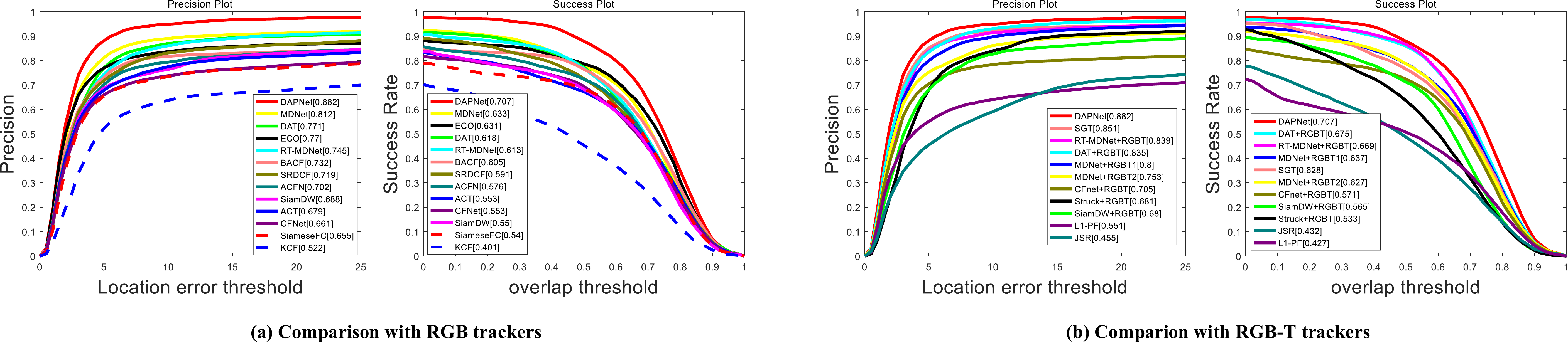} \\
  \caption{Evaluation curves on GTOT dataset. The representative scores of PR/SR are presented in the legend. For clarity, we separate RGB and RGBT trackers in (a) and (b) respectively. }\label{fig::Curve-GTOT}
\end{figure*}

\section{Experiments}

We evaluate our method on two popular RGBT tracking benchmark datasets, GTOT ~\cite{Li16tip} and RGBT234 ~\cite{li2019rgb}, and compare with existing trackers. 
The experiments are conducted on the following specifications, 4.2 GB Intel core i7-7700K CPU, 32 GB RAM, and
NVIDIA GeForce GTX 1080Ti GPU using PyTorch.

\begin{figure*}[t]
  \centering
  \includegraphics[width=\textwidth]{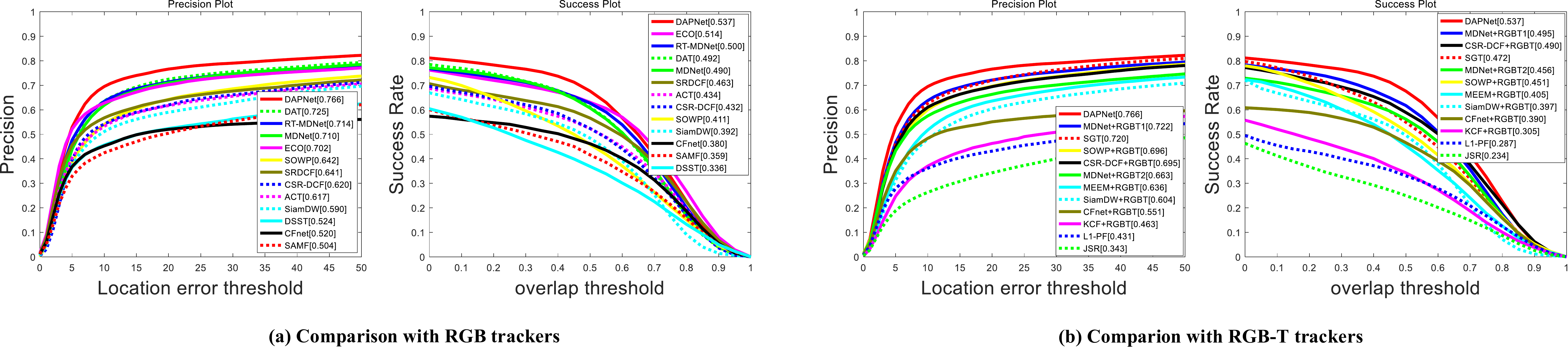} \\
  \caption{Evaluation curves on RGBT234 dataset. The representative scores of PR/SR are presented in the legend. For clarity, we separate RGB and RGBT trackers in (a) and (b) respectively. }\label{fig::Curve-RGBT234}
\end{figure*}

\subsection{Evaluation Setting}

{\flushleft \bf Datasets.}
There are only two large RGBT tracking datasets, i.e., GTOT ~\cite{Li16tip} and  RGBT234 ~\cite{li2019rgb}. 
They are large and challenging enough, and we evaluate our approach on them for comprehensive validation.  
GTOT includes 50 aligned RGBT video pairs with about 15K frames in total. 
And each frame pair is annotated with ground truth bounding box.  
RGBT234 dataset is a large-scale RGBT tracking dataset extended from RGBT210 dataset~\cite{Li17rgbt210}. 
It contains 234 RGBT videos and each video has a RGB video and a thermal video. 
Its total number of frames reach about 234,000 and the number frames of the longest video pair reaches 8,000. 
To analyze the attribute-based performance of different tracking algorithms, it is annotated with 12 attributes.

{\flushleft \bf Parameters.}
Each sample is resized to $107 \times 107$ as the input to the network. 
When training the network, we set the learning rate of the convolutional layer and the fully connected layer to 0.0001, 0.001 respectively. 
In the tracking phase, the convolutional layer is not updated, the learning rate of fc4-fc5 is 0.0001, and the learning rate of fc6 is 0.001. 
The weight decay and momentum are set to 0.0005 and 0.9, respectively. 
At the initial frame of a test sequence, we train the fully connected layers for 10 iterations. 
In particular, we set a gradient threshold $clip \_ gradient$ is 100.

{\flushleft \bf Evaluation metrics.}
 On these two datasets, we utilize two widely used metrics, precision rate (PR) and success rate (SR), to evaluate RGBT tracking performance.
PR is the percentage of frames whose output location is within the given threshold distance of groundtruth.  
We set the threshold to be 5 and 20 pixels for GTOT and RGBT234 datasets respectively (since the target object in the GTOT dataset is generally small) to obtain the representative PR. 
Similarly, SR is the ratio of the number of successful frames whose overlap is larger than a threshold. 
By varying the threshold, the SR plot can be obtained,
and we employ the area under curve of SR plot to define the representative SR.

{\flushleft \bf Baseline.}
Our baseline method is MDNet, which is a RGB tracking algorithm. 
For a more fair comparison, we extend MDNet algorithm for RGBT tracking of dual-modal inputs. 
In our experiment, we extend MDNet to two forms.  
One is that we directly concatenate two modes of data channels to form 6 channels of input data, and then input the original MDNet algorithm for tracking, which is named MDNet+RGBT1. 
The other is that we extract the convolution features of the two modes separately, and concatenate the feature maps of the two modes at conv3, named MDNet+RGBT2. 
From the experimental results in Table ~\ref{tb::baseline}, it can be seen that the MDNet+RGBT2 is significantly worse than MDNet and MDNet+RGBT1 is not significantly superior to MDNet either, even PR scores are lower than MDNet. 
There are two possible reasons. 1) Directly concatenating two modal data does not make effective use of complementary information between modalities. 
2) Redundant features and noise interference might be introduced. 
These validate that the direct and simple concatenate of two modal data does not achieve good tracking performance, and also verify the effectiveness of our dense feature aggregation and pruning.

\begin{table}[t]\footnotesize
\caption{Performance of our method against MDNet with different implementation strategies on GTOT and RGBT234 datasets.}
\centering
\begin{tabular}{c| c  |  c   c  c |  c }
	\hline
   & & MDNet  & MDNet+RGBT1  &MDNet+RGBT2  & DAPNet\\\hline
 GTOT&PR & 81.2 & 80.0 & 75.3 & 88.2 \\
 &SR & 63.3 & 63.7 & 62.7 & 70.7 \\\hline

RGBT234&PR &71.0 & 72.2 &66.3 & 76.6 \\
 &SR & 49.0 &49.5 &45.6 & 53.7 \\\hline
\end{tabular}
\label{tb::baseline}
\end{table}

\subsection{Evaluation on GTOT}
{\bf Comparison with RGB trackers.}
To verify the superiority of the proposed RGBT tracking method compared to RGB tracking, we first evaluate our method with twelve state-of-the-art RGB tracker, including DAT~\cite{Pu2018Deep}, RT-MDNet~\cite{Jung2018Real}, SiamDW~\cite{Zhipeng2019Deeper}, ACT~\cite{chen2018real}, MDNet~\cite{MDNet15cvpr}, ECO~\cite{ECO17cvpr}, BACF~\cite{Galoogahi2017Learning} SRDCF~\cite{danelljan2015learning}, ACFN~\cite{Choi2017Attentional}, SiameseFC~\cite{Bertinetto2016Fully}, CFnet~\cite{CFNet17cvpr} and KCF~\cite{henriques2015high}. 
Figure ~\ref{fig::Curve-GTOT}(a) shows that our method outperforms these trackers, demonstrating the effectiveness of introducing thermal information in visual tracking. 
In particular, our tracker outperforms MDNet and DAT with 7.0\%/7.4\%, 11.1\%/8.9\% in PR/SR, respectively.

{\bf Comparison with RGBT trackers.} 
We further compare our tracker with several state-of-the-art RGBT trackers, including RT-MDNet+RGBT, DAT+RGBT, SiamDW+RGBT, CSR~\cite{Li16tip}, JSR~\cite{Sun2012Fusion}, L1-PF~\cite{Wu11icif}, SGT~\cite{Li17rgbt210}, MDNet+RGBT1, and MDNet+RGBT2.  
Since there are few RGBT trackers~\cite{Li16mmm,Liu12infosci,Li2016Learning,Li17rgbt210}, some RGB tracking methods have been extended to RGB-T ones by concatenating RGB and thermal features into a single vector or regarding the thermal as an extra channel, such as RT-MDNet, DAT, SiamDW and CFnet. 
Figure ~\ref{fig::Curve-GTOT}(b) shows that our tracker significantly outperforms them, demonstrating the effectiveness of employing RGB and thermal information adaptively to construct robust feature representations in our approach. 
In particular, our tracker achieves 3.1\%/7.9\% and 4.3\%/3.8\% performance gains in PR/SR over SGT and RT-MDNet+RGBT.

\subsection{Evaluation on RGBT234}
For more comprehensive evaluation, we report the evaluation results on the RGBT234 dataset ~\cite{li2019rgb}, as shown in Figure~\ref{fig::Curve-RGBT234}. 
The comparison trackers include twelve RGB ones (DAT~\cite{Pu2018Deep}, RT-MDNet~\cite{Jung2018Real}, SiamDW~\cite{Zhipeng2019Deeper}, ACT~\cite{chen2018real}, MDNet~\cite{MDNet15cvpr}, ECO~\cite{ECO17cvpr}, SOWP~\cite{Kim15iccv}, SRDCF~\cite{danelljan2015learning}, CSR-DCF~\cite{dcf-csr16cvpr}, DSST~\cite{danelljan2014accurate}, CFnet~\cite{CFNet17cvpr} and SAMF~\cite{li2014scale}) and eleven RGBT ones (MDNet+RGBT1, MDNet+RGBT2, SGT~\cite{Li17rgbt210}, SOWP+RGBT, CSR-DCF+RGBT, MEEM~\cite{MEEM14eccv}+RGBT, CFnet+RGBT, KCF~\cite{henriques2015high}+RGBT, JSR~\cite{Sun2012Fusion} and L1-PF~\cite{Wu11icif}). 
From the results, we can see that the performance of our method clearly outperforms the state-of-the-art RGB and RGBT methods in all metrics. 
It demonstrates the importance of thermal information and our method. 
In particular, our method achieves 4.1\%/4.5\% performance gains in PR/SR over the second best RGB tracker DAT, and achieves 4.4\%/4.2\% and 4.6\%/6.5\% gains over MDNet+RGBT1 and SGT, respectively. 
Note that the methods based on Siamese network~\cite{chopra2005learning} have weak performance on RGBT234 and GTOT, including SiamDW, CFNet and SiameseFC. 
It is because of the great gap between the training datasets and the test datasets (RGBT234 and GTOT), and those methods might only fit the training dataset very well but the test results are poor.

\begin{table}[t]\footnotesize
\caption{Evaluation results of our DAPNet with its variants on RGBT234 dataset.}
\centering
\begin{tabular}{c| c  |  c   c  c |  c }
	\hline
   & & DAPNet-noFACP  &  DAPNet-noFA  & DAPNet-noFP  &  DAPNet\\\hline
 RGBT234&PR & 66.3 & 67.0 & 74.3 & 76.6 \\
 &SR & 45.6 & 47.1 & 52.5 & 53.7 \\\hline

\end{tabular}
\label{tb::componet-analysis}
\end{table}

\begin{figure*}[t]
  \centering
  \includegraphics[width=\textwidth]{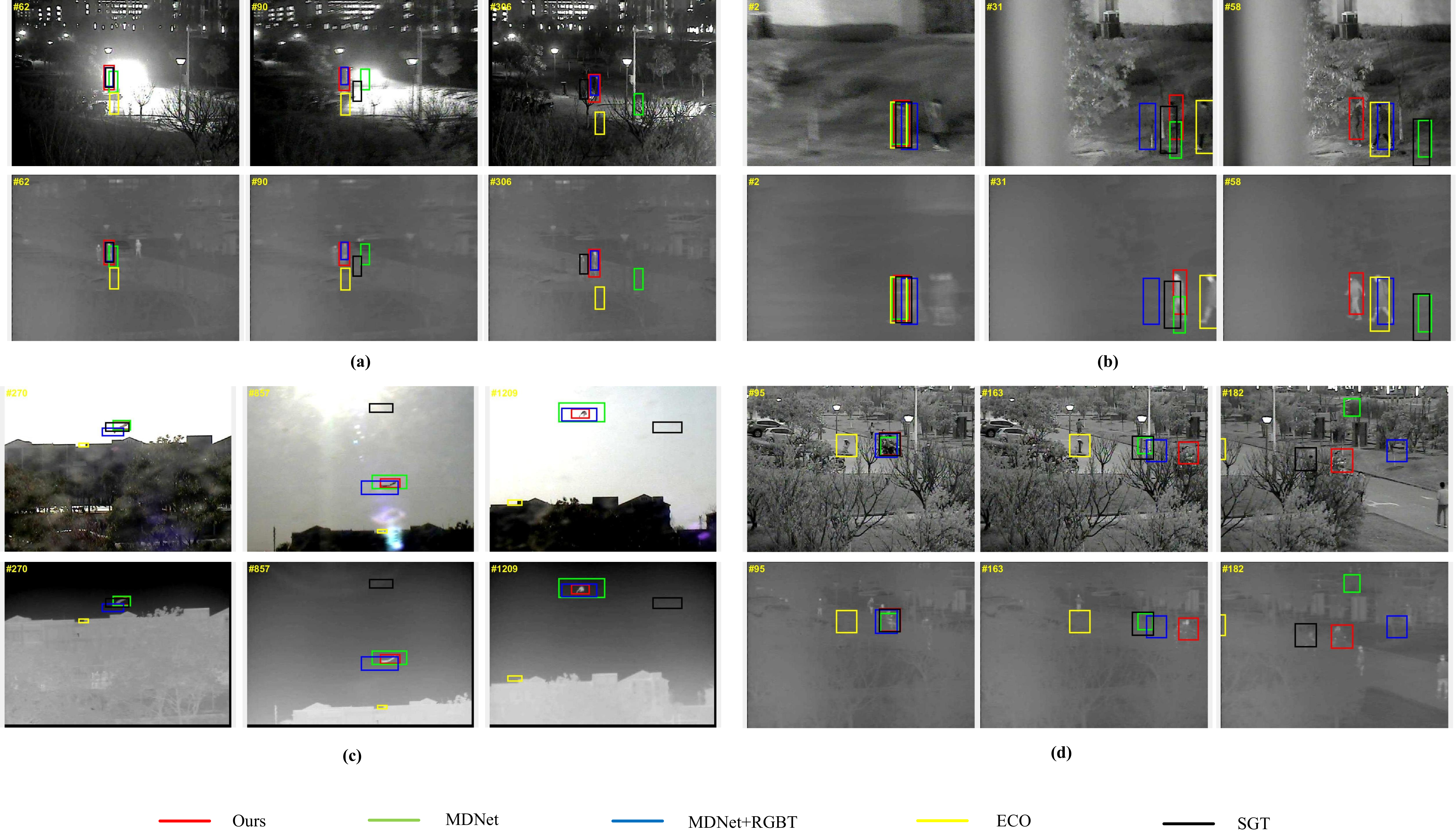} \\
  \caption{ Visual comparison of our DAPNet versus four state-of-the-art trackers on four video sequences. }\label{fig::sample}
\end{figure*}

{\bf Attribute-based performance.} 
We also report the attribute-based results of our method versus other state-of-the-art RGBT trackers on RGBT234 dataset, including L1-PF, KCF+RGBT, MEEM+RGBT, SOWP+RGBT, CSR-DCF+RGBT, CFNet+RGBT, SGT and MDNet+RGBT1, as shown in Table ~\ref{tb::AttributeResults}. 
The attributes include no occlusion (NO), partial occlusion (PO), heavy occlusion (HO), low illumination (LI), low resolution (LR), thermal crossover (TC), deformation (DEF), fast motion (FM), scale variation (SV), motion blur (MB), camera moving (CM) and background clutter (BC). 
The results show that our method performs the best in terms of most challenges except for NO, PO, LI, BC, and SV. 
It demonstrates the effectiveness of our method in handling the sequences with the appearance changes and adverse conditions.

\subsection{Ablation Study}
To justify the significance of the main components using RGBT234, we implement 3 special versions of our approach for comparative analysis including, 
1)  DAPNet-noFACP, that removes the dense feature aggregation and feature pruning, which is {MDNet+RGBT2}. 
2) DAPNet-noFA, that removes the dense feature aggregation and only use feature pruning. 
3) DAPNet-noFP, that removes the feature pruning and only use dense feature aggregation. 
From Table ~\ref{tb::componet-analysis} we can see that, 
1) Introducing the dense feature aggregation has a big performance boost by observing  DAPNet-noFP against DAPNet-noFACP.
2) DAPNet-noFA is better than DAPNet-noFACP, and the similar observation can be obtained for DAPNet with DAPNet-noFP.
It suggests that the feature pruning scheme could handle noise interference, and thus lead to tracking performance improvement. 
3) DAPNet outperforms DAPNet-noFACP,  DAPNet-noFA and  DAPNet-noFP, which justify the effectiveness of the whole framework with the dense feature aggregation and feature pruning.

\subsection{Qualitative Performance}
The qualitative comparison of our algorithm versus two state-of-the-art RGB trackers and two state-of-the-art RGBT trackers on partial video sequences is presented in Figure~\ref{fig::sample}, including MDNet ~\cite{Nam2015Learning}, ECO~\cite{ECO17cvpr}, MDNet+RGBT2 and SGT ~\cite{Li17rgbt210}. 
Our approach shows consistently better performance in various challenging scenarios including high illumination, motion blur, scale variation and background clutter. 
For example, in Figure~\ref{fig::sample} (a), the \emph{man} is totally invisible in RGB source but the thermal images can provide reliable information to distinguish the target from the background. 
In Figure~\ref{fig::sample}(d), our method performs well in presence of partial occlusions and background clutter while other trackers lose the target. 
Compared with four trackers, the superiority of our tracker is fully demonstrated  in these extreme challenges. 
It also verifies that dense feature aggregation and  pruning can enhance image features to obtain better tracking results.

\section{Conclusion}
We have proposed an end-to-end deep network for RGBT tracking.  
Our network consists of two major modules, one is dense feature aggregation that provides a powerful RGBT feature repsentations for target objects, and another is feature pruning that collaboratively select most discriminative feature maps from two modalities for enhancement of RGBT features. 
Extensive experiments on two benchmark datasets demonstrate that our approach is able to handle various challenges like background clutter and partial occlusion, and thus improves tracking performance significantly. 
In future work, we will investigate a wider and deeper network~\cite{Zhipeng2019Deeper} in our framework to further boost RGBT tracking performance, and improve the network structure (i.e.,~\cite{Jung2018Real} extract more accurate representations of targets and candidates by RoIAlign ) to achieve real-time performance.

\begin{acks}
This research is jointly supported by the National Natural Science Foundation of China (No. 61860206004, 61702002, 61872005), Natural Science Foundation of Anhui Province (1808085QF187), Open fund for Discipline Construction, Institute of Physical Science and Information Technology, Anhui University.

\end{acks}

\bibliographystyle{ACM-Reference-Format}
\bibliography{sample-base}


\end{document}